\documentclass[letterpaper, 10 pt, conference]{ieeeconf}

\IEEEoverridecommandlockouts
\usepackage{cite}

\usepackage[pagewise]{lineno}
\usepackage{makecell}
\usepackage{threeparttable}
\usepackage{graphicx}
\usepackage{color}
\usepackage{algorithm}
\usepackage{algpseudocode}
\usepackage{amsmath}
\usepackage{diagbox}
\usepackage{multirow}

\graphicspath{{./figures/}}

\title{\LARGE \bf
Toward Efficient and Robust Multiple Camera Visual-inertial Odometry
}

\author{Yao He$^{1, 2}$, Huai Yu$^{2}$, Wen Yang$^{3}$ and Sebastian Scherer$^{2}$% <-this % stops a space
\thanks{$^{1}$Yao He is with the School of Science and Engineering, The Chinese University of Hongkong, Shenzhen and Shenzhen Institute of Artificial Intelligence and Robotics for Sociaty, Shenzhen, China. {\tt\small yaohe@link.cuhk.edu.cn}}%
\thanks{$^{2}$Huai Yu, and Sebastian Scherer is with the AirLab, Carnegie Mellon University, Pittsburgh, PA 15213, USA. {\tt\small \{huaiy,basti\}@andrew.cmu.edu}}%
\thanks{$^{3}$Wen Yang is with the Electronic Information School, Wuhan University,  Wuhan 430072, China. {\tt\small yangwen@whu.edu.cn}}%
}

\begin{document}
% \linenumbers % Uncomment this to enable line numbers in the peer review

\maketitle
\thispagestyle{empty}
\pagestyle{empty}

%%%%%%%%%%%%%%%%%%%%%%%%%%%%%%%%%%%%%%%%%%%%%%%%%%%%%%%%%%%%%%%%%%%%%%%%%%%%%%%%
\begin{abstract}
Efficiency and robustness are the essential criteria for the visual-inertial odometry (VIO) system. To process massive visual data, the high cost on CPU resources and computation latency limits VIO’s possibility in integration with other applications. Recently, the powerful embedded GPUs have great potentials to improve the front-end image processing capability. Meanwhile, multi-camera systems can increase the visual constraints for back-end optimization. Inspired by these insights, we incorporate the GPU-enhanced algorithms in the field of VIO and thus propose a new front-end with NVIDIA Vision Programming Interface (VPI). This new front-end then enables multi-camera VIO feature association and provides more stable back-end pose optimization. Experiments with our new front-end on monocular datasets show the CPU resource occupation rate and computational latency are reduced by 40.4\% and 50.6\% without losing accuracy compared with the original VIO. The multi-camera system shows a higher VIO initialization success rate and better robustness overall state estimation.
\end{abstract}

%%%%%%%%%%%%%%%%%%%%%%%%%%%%%%%%%%%%%%%%%%%%%%%%%%%%%%%%%%%%%%%%%%%%%%%%%%%%%%%%
\section{INTRODUCTION}

Visual-inertial state estimation serves as the most fundamental role for a wide range of applications, such as robotic navigation, autonomous driving, and augmented reality (AR) \cite{QinTRO2018}. Current monocular and stereo Visual-inertial Odometry systems have achieved great success for state estimation \cite{QinTRO2018,BloeschIJRR2017,murORB2}. Generally, a VIO algorithm composes of a front-end to process image data and a back-end that carries out the optimization. 
However, because of the substantial visual data stream, the front-end usually takes extensive CPU resources, yielding the computation latency and incompatibility with other tasks on a computation-limited chip. Additionally, monocular and stereo VIO sensors have limited field-of-view (FoV), making the state output unstable when the observed scene is homogeneous. These two main challenges make the state-of-the-arts far away from real applications.
   \begin{figure}[thpb]
      \centering
      \includegraphics[scale=0.29]{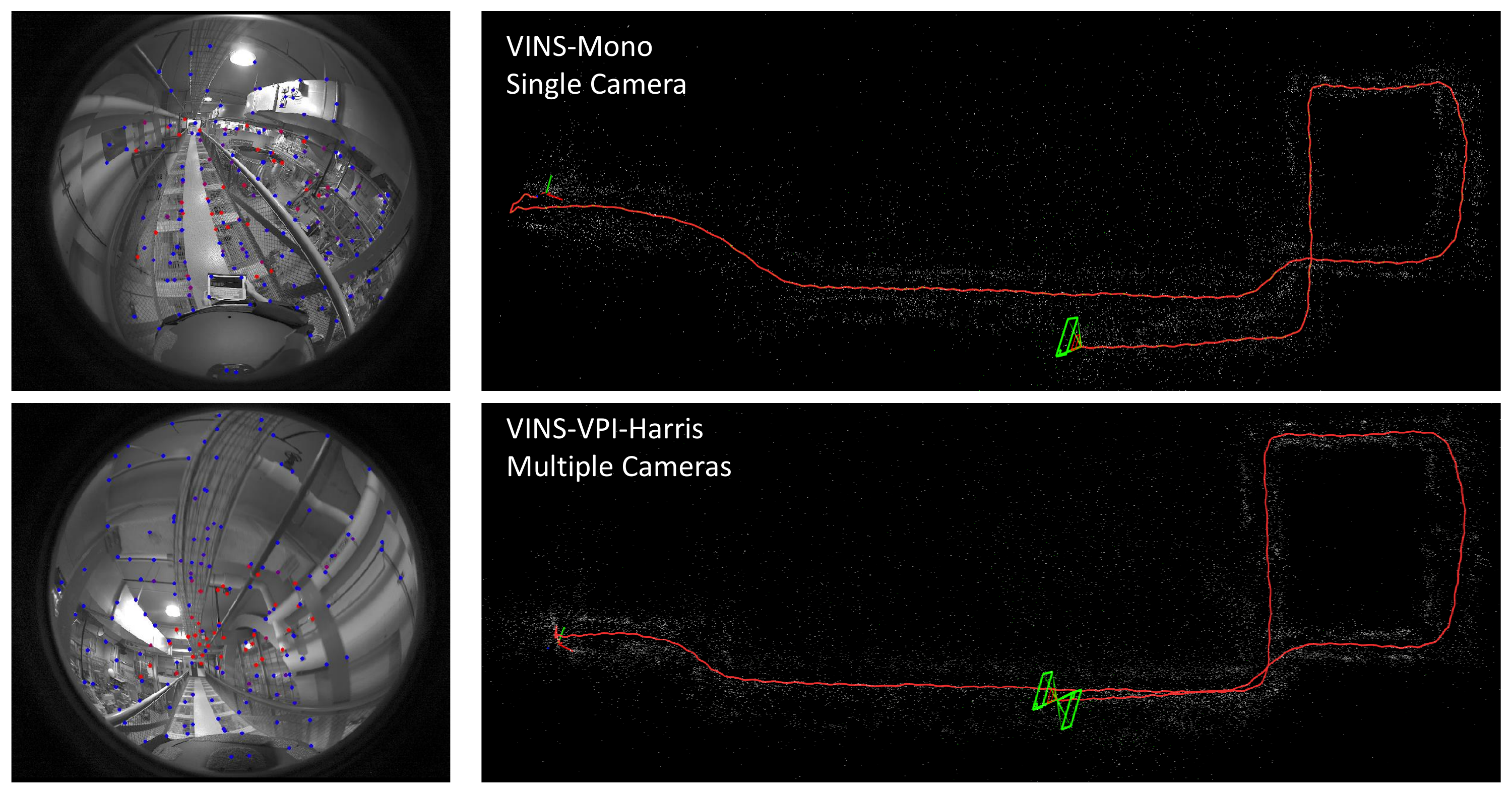}
      \caption{\emph{Left (top-bottom):} two non-overlapped images of multi-camera system and the tracked features (red). \emph{Right-top:} the trajectory of original VINS-Mono \cite{QinTRO2018} on a single camera. \emph{Right-bottom:} the trajectory of our proposed efficient multi-camera VIO system. The multi-camera system shows a much smaller drift when back to the line around a rectangle loop (loop closure is disabled for both methods).}
      \label{title}
      \vspace{-0.5cm}
   \end{figure}
% first challenge
The VIO front-end is usually used to extract landmark features and track features among video frames. To ensure the stability of feature tracking, the input videos are often at a high frequency above 20 Hz. Consequently, it takes a large proportion of CPU resources to guarantee real-time performance. Jeon \cite{jeon2021run} tests the CPU usage of different VIO algorithms on various NVIDIA hardware (Jetson TX2, Xavier NX, and AGX Xavier boards). His work shows that most of the VIO algorithms have extensive CPU usages around 150\%$\sim$250\%.  Besides, computation latency resulted from the high video frame rate and limited CPU capability is a significant problem for real-time VIO systems. The accumulating delay will get worse when the system operates for a long time. To overcome the computation latency, the SOTA methods \cite{QinTRO2018, BloeschIJRR2017, murORB2} limit the number of tracked features. However, the CPU occupation remains high. In this work, we build a new front-end to release the CPU usage and the computation latency while the estimation accuracy maintains or even gets improved. 

% second challenge, limited FoV, unstable results, mult camera system,
Robustness is another vital metric for VIO systems. Although the current SOTA monocular and stereo VIO systems have achieved stable performance on most public datasets, the limited scenarios and motion patterns in existing public datasets \cite{tartanair2020iros} making such stability unreliable. In typical cases, such as when the cameras face close to objects, the limited FoV of monocular and stereo cameras results in significant occlusion, making the feature tracking performance unstable. 

To account for this problem, numerous works \cite{oskiper2007visual, kazik2012real, xiang2019vilivo} have been proposed to use multiple cameras to increase the FoV. Thus the most stable tracking features can be selected among these cameras. However, the computational burden will increase significantly as the number of cameras increases. This kind of computation occupation prevents other applications, such as object detection and motion planning, from running on the same computer. Therefore, there is a great need to enable a multi-camera VIO system to handle massive visual data at a low CPU usage. 
% new para, our contribution.

To address the above two problems when improving the efficiency and robustness of VIO systems, we mainly focus on incorporating the graphics processing units (GPUs) enhanced modules into the VIO algorithms. Recently, the powerful embedded GPUs have been utilized as the front-end of VIO to improve the information processing capability \cite{Nagy2020}. In this paper, we first incorporate the Vision Programming Interface (VPI) toolbox \cite{VPI} to propose a new GPU accelerated VIO front-end, which significantly reduces the CPU usage and maintains the state estimation efficiency and accuracy. Then we further extend the monocular VIO to a multi-camera VIO system, where cameras have non-overlapping FoV. All the front-end tasks are conducted on GPU with VPI, and all the features are integrated into one back-end with IMU measurements. The configuration of multi-camera VIO finally improves the initialization success rate and robustness on extensive evaluation with our collected datasets. The major contributions are listed as follows:
\begin{itemize}
    \item We propose a GPU enhanced front-end implemented by VPI for low cost on computational latency and CPU usage while maintaining state estimation accuracy and robustness.
    \item With the efficient front-end, we extend the monocular pipeline to a  multi-camera VIO system. All the features observed in multiple cameras are fed into the back-end for optimization, which provides a higher system initialization success rate and more robust state estimation.
\end{itemize}
% \item Images from multiple cameras are all stored into previous map database.  

The rest of this paper is organized as follows. Section II provides the related literature. Section III describes the VPI accelerated front-end and the multi-camera VIO system. Section IV provides the experimental results and discussions. Finally, this paper is concluded in Section V.

   \begin{figure*}[!t]
      \centering
      \includegraphics[scale=0.40]{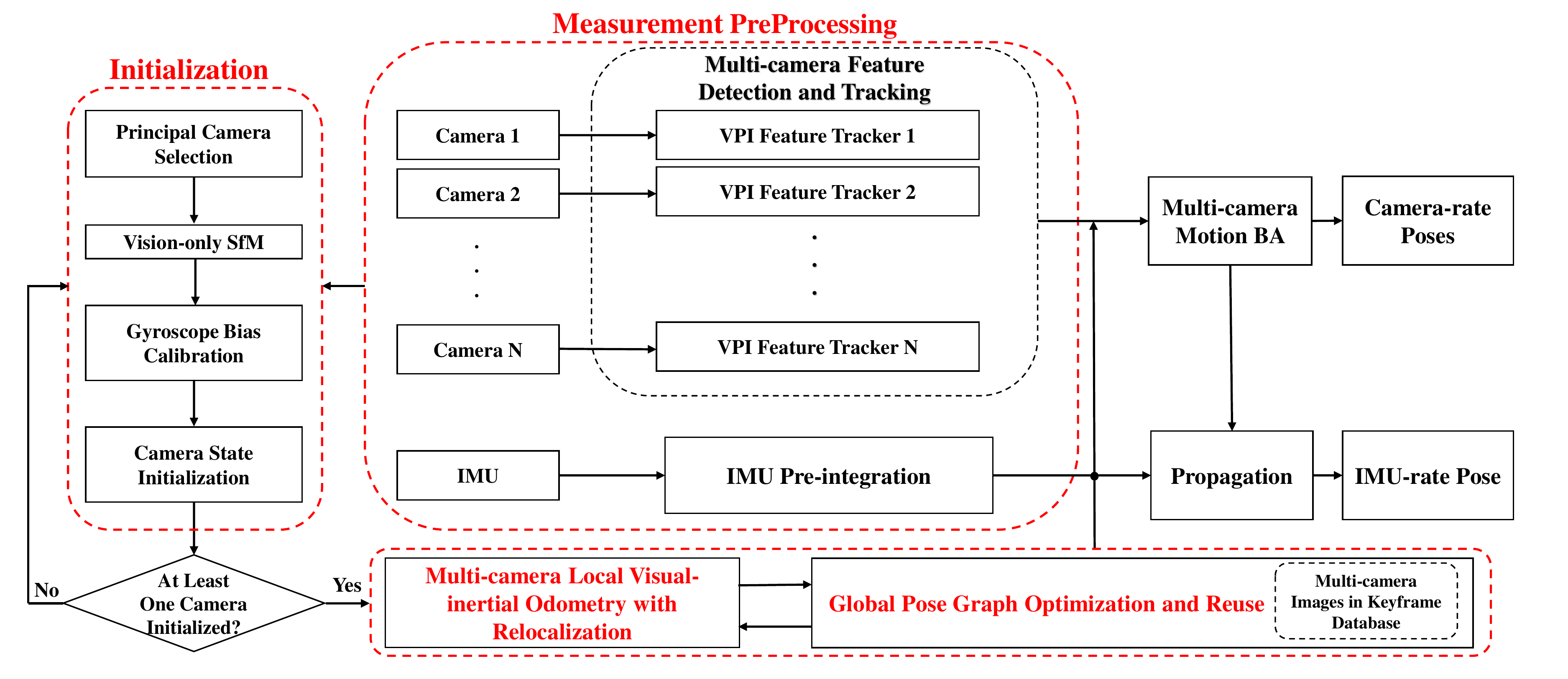}
      \caption{Block diagram illustrating the full pipeline of the proposed multi-camera VIO with VPI enhanced front-end.}
      %The diagram describe the differences with the VINS-Mono in details, while  omit details in sections that are kept almost the same. The blocks with blue edges are the sections with few modification.}
      \label{system}
      \vspace{-0.5cm}
   \end{figure*}
\section{RELATED WORK}

\subsection{GPU Accelerated VIO Algorithms}

Numerous VIO algorithms have been proposed in recent years, such as VINS-Mono \cite{QinTRO2018}, MSCKF \cite{mourikis2007multi}, ROVIO \cite{BloeschIJRR2017} and  ORB-SLAM3 \cite{campos2021orb}. Most of them consist of a front-end for feature association and a back-end for pose/map optimization. Although these VIO methods have achieved good results for online state estimation, The CPU usages are all high because massive visual images need to be processed. According to the experiments conducted on Xavier \cite{jeon2021run}, the CPU usage of VINS-Mono is about 150-170$\%$ on multi-core processing, MSCKF is above 170\%. ROVIO has the least CPU usages (around 60\%), but the back-end EKF is less efficient compared to the bundle adjustment method \cite{c9}. The computational latency on the back-end will be larger in ROVIO. ORB-SLAM has the largest CPU usage ranging about 150-240$\%$. These large CPU occupations make the VIO systems cannot run on the same computer when robots have other tasks, such as object detection and motion planning. Additionally, the limitation to the CPU usage will increase the computational latency. Therefore, the trade-off between CPU usage and efficiency for VIO systems has always been a difficult problem. 

With the hardware revolution in the last decade, GPUs are installed in embedded platforms, such as NVIDIA Xiaver, which enables more parallel computing capacity. Recently, several works have utilized GPUs to improve the VIO front-end video processing capability and reduce latency. The commonly used feature detectors, such as Shi-Tomasi \cite{shi1994good}, Harris \cite{harris1988combined},  and FAST \cite{rosten2006machine} are implemented in CUDA Visual Library (VILIB) \cite{Nagy2020} for fast feature extraction. VILIB applies efficient low-level, GPU hardware-specific programming for feature detection and feature tracking. Recently, NVIDIA proposed the Vision Programming Interface (VPI) \cite{VPI}. VPI enables lots of basic computer vision algorithms (such as Harris detector and LK optical flow) on either CPU, GPU, VPA ( Programmable Vision Accelerator), and VIC (Video Image Compositor). It supports an easy switch between different processing chips. These GPU implementations have great potentials to shift the computational-expensive VIO front-end to GPU, reducing the CPU usage and computation latency.

\subsection{Multiple Camera Visual Odometry}
Although the SOTA monocular and stereo VIO systems have impressive performance, they are still not robust enough for autonomous state estimation on real robots over a long period with complex motions and complicated environments. Multiple cameras can significantly add visual constraints with wider FoV and thus provide redundancy for back-end optimization. Taragay \emph{et al.} proposed the loosely-coupled VIO system on two pairs of backward and forward stereo cameras with an IMU \cite{oskiper2007visual}. Tim \emph{et al.} introduced a real-time 6D stereo visual odometry with two non-overlapping FoV cameras \cite{kazik2012real}. Liu \emph{et al.} presented a pose tracker and a local mapper supporting an arbitrary number of stereo cameras \cite{liu2018towards}. These methods demonstrate the robustness of multi-camera visual odometry over a long traveling distance and complex environments. However, all the multi-camera processing is based on the multi-core CPU parallel computing, adding great computation burden to CPU and inevitably bringing computation latency to the state estimation. With the GPU accelerated front-end mentioned before, our goal is to set the front-ends of multiple cameras conducted on GPU without consuming any CPU resources so that the back-end optimization can have enough cores to ensure non-latency estimation.  

% \subsection{Benchmark Comparison of VIO}
% Numerous studies have been conducted on the benchmarking of VO or VIO methods. The EuRoC datasets \cite{Burri25012016} are commonly used in different VIO works, such as \cite{QinTRO2018}, \cite{jeon2021run}, and \cite{Nagy2020}. These datasets contain stereo images, synchronized IMU measurements, and accurate motion and structure ground-truth. They can be used to test both mono and stereo VIOs. EuRoC datasets provides three difficulty levels and three different scenes. Such diversity allows researchers to test VIO algorithms in different environments and motions. The rpg\_trajectory\_evaluation repository \cite{Zhang18iros} implements common used trajectory evaluation methods for VO/VIOs. This repository supports both single and multiple trajectory estimation, as well as comparison between different algorithms on many datasets, including EuRoC MAV Datasets. 

\section{Methodology}
\subsection{Overview}
The structure of the proposed multi-camera visual-inertial state estimator is shown in Fig. \ref{system}. We first extract and track visual features from non-overlapping cameras individually with VPI tools. IMU measurements between two consecutive frames are preintegrated as in VINS-Mono \cite{QinTRO2018}. The initialization procedure selects one principal camera to initialize and ends with initialization on all the subordinate cameras. It is regarded as a  success if at least one camera is initialized successfully. All the cameras' features will be included in the visual-inertial bundle adjustment. In terms of relocalization and pose graph, all the initialized cameras' frames will be stored in the keyframe database for loop detection. 
      
The notations and frame definitions that we use throughout the paper are defined as follows. $(\cdot)^{c_{i}}$, $(\cdot)^{b}$, and $(\cdot)^{w}$ represent the \textit{i}-th camera frame, body (IMU) frame and world frame, respectively. Rotation matrices $\boldsymbol{R}$ and Hamilton quaternions $\boldsymbol{q}$ both represent rotation. $\otimes$ represents the multiplication operation between two quaternions. In our system, we assume that the IMU and cameras are already calibrated. Thus the extrinsic parameters are treated as known quantities.

\subsection{VPI-enhanced Front-end}
\begin{figure}[htpb]
   \vspace{-0.3cm}
      \centering
      \includegraphics[scale=0.43]{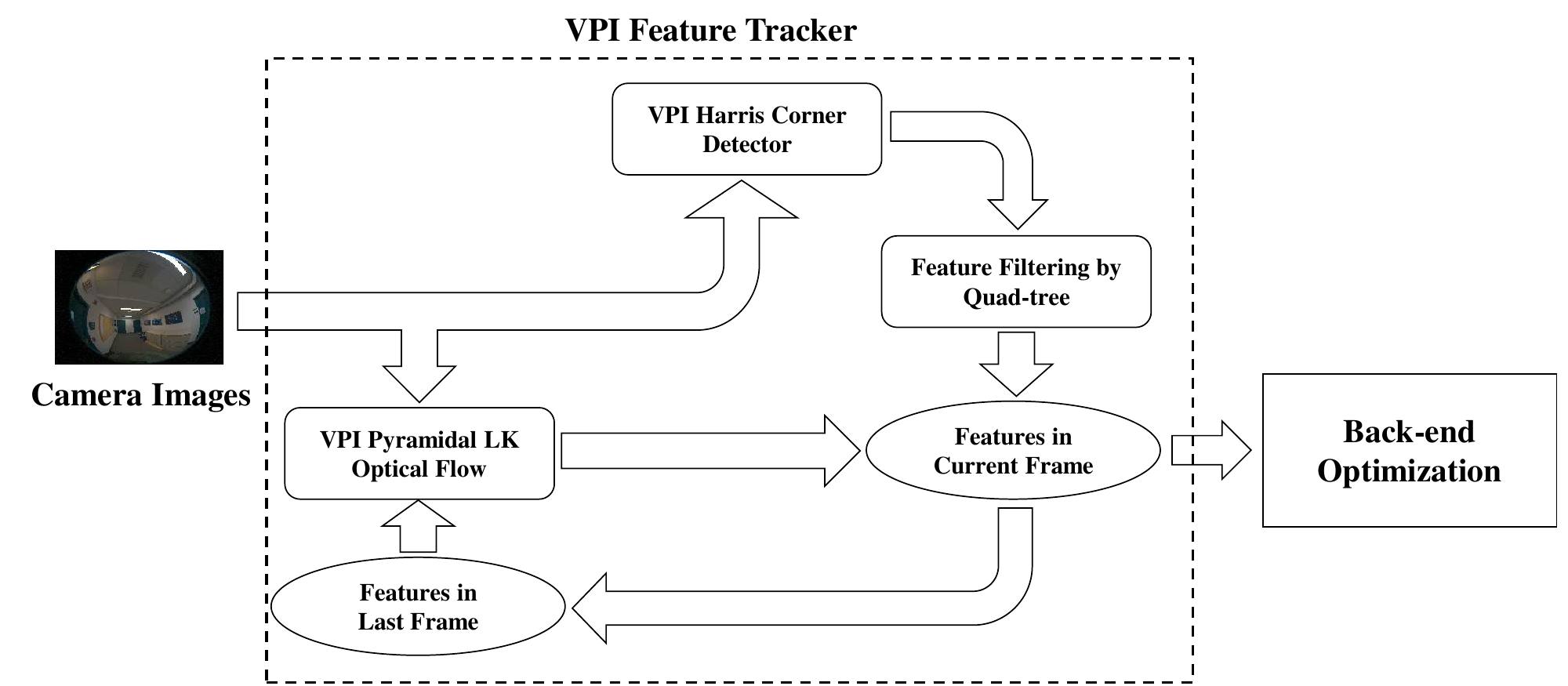}
      \caption{The overall front-end architecture of VINS-Mono with VPI.}
    %   The VPI Harris Corner Detector detects new feature points. The VPI Pyramidal LK Optical Flow tracks the feature points. A filter implemented by quad-tree algorithm makes the distribution of feature points more uniform. Both the tracked points and detect points will be passed to the back-end for further processing
      \label{architecture}
      \vspace{-0.5cm}
   \end{figure}

The front-end of a VIO detects and tracks features, and passes the visual information to the back-end for optimization. VPI provides Harris corner detector to detect features and pyramidal LK optical flow to track features. The overall architecture of the front-end is demonstrated in Fig. \ref{architecture}. 

   \begin{figure}[htpb]
      \centering
      \includegraphics[scale=0.43]{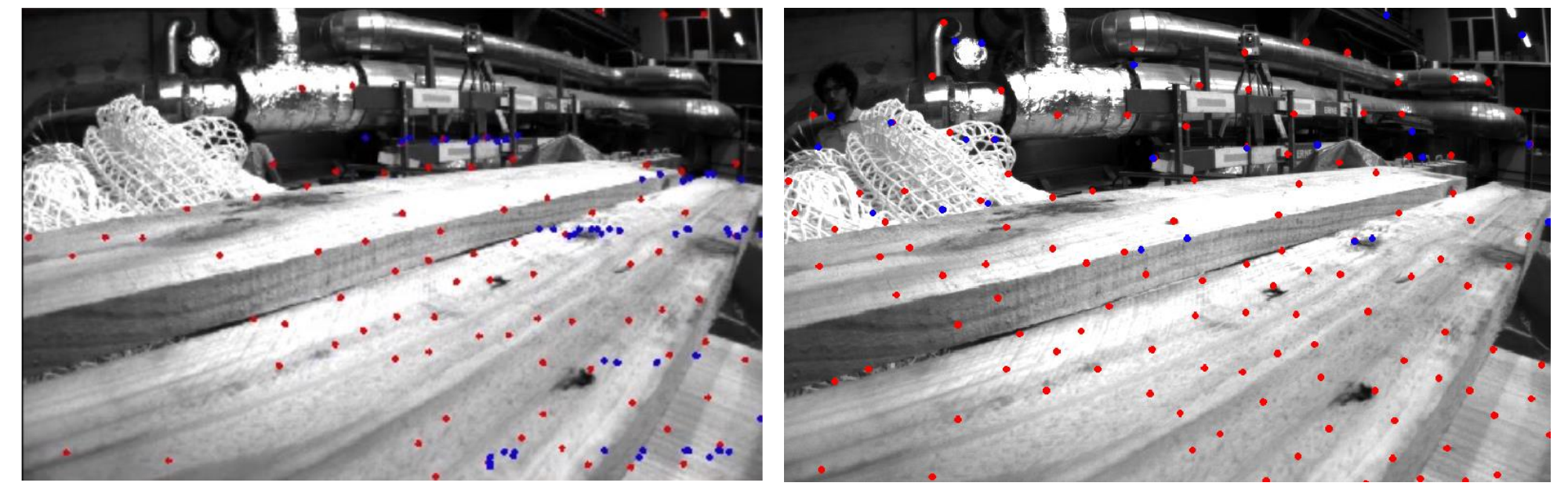}
      \caption{Distribution of feature points. The scene is from EuRoC dataset \cite{Burri25012016}. \textit{Picture on the left}: feature points picked by their scores. \textit{Picture on the right}: feature points after the quad-tree filter. \textit{Red points}: tracked features. \textit{Blue points}: newly detected features.}
      \label{quadTree}
      \vspace{-0.3cm}
   \end{figure}
   Image stream will be directly transferred to GPU, and we use the Harris corner detector in the VPI toolbox to detect features. To maintain the feature number and feature continuity between frames, features are ranked by the response scores.  However, if the Harris features are picked directly according to their scores, their distribution will be concentrated in richly textured areas. This reduces the robustness in initializing the VIO and state estimation. Inspired by ORB-SLAM \cite{murORB2}, we propose to use the quad-tree algorithm to filter the feature points. The point with the highest response score will be selected regardless of how many points in a local patch.  It ensures a more uniform distribution of feature points, as shown in Fig. \ref{quadTree}. The incoming image will also be transferred to pyramidal LK optical flow to track the feature points in the last frame. The newly detected feature points and the tracked points constitute the feature points in the current image. They will be passed to the back-end for further processing. Data storage and information transportation are purely based on VPI defined data structures, which allow minimum cost on transporting data between CPU and GPU. 
   
% One major property of VPI is to support easy interoperation with existing projects that make use of OpenCV libraries. Therefore, the feature detection and tracking modules are directly substituted with corresponding VPI equivalents (Harris Corner Detector and Pyramidal LK Optical Flow). 

% \begin{algorithm}[htb]
%   \caption{ Selecting Feature points by Quad-tree.}
%   \label{alg:Framwork}
%   \begin{algorithmic}[1]
%     \Require
%       $P_{in}$: the set of detected feature points; 
%       $S$: the set of scores corresponding to detected feature points;
%       $k$: the number of needed feature points;
%       $M$: mask of fisheye.
%     \Ensure
%          $P_{out}$: the set of selected feature points;
%     \State Select the feature points inside $M$ into set $P_{in}'$;
%     \If{$|P_{in}'|<n$} 
%         \State \Return $P_{in}'$;
%     \EndIf
%     \State Initialize image node set $N$, first image node $n_1$;
%     \State Store all feature points in $P_{in}'$ into $n_1$;
%     \State Add $n_1$ into $N$;
%     \While{$|N|<k$}
%         \For{$n_i \in N $}
%             \If{$n_i$ has no feature point}
%                 \State Remove $n_i$ from $N$
%             \EndIf
%             \If {$n_i$ has at least 2 feature point}
%                 \State Split $n_i$ to four sub-nodes with the same image size;
%             \EndIf

%         \EndFor
%     \EndWhile
%     \For {$n_i$ in $N$}
%         \State Select the feature point $p_i$ having the highest score;
%         \State Add $p_i$ to $P_{out}$;
%     \EndFor
%     \State \Return $P_{out}$;
%   \end{algorithmic}
% \end{algorithm}

\subsection{Estimator Initialization}
Since the cameras have non-overlapping FoV, the system needs an accurate scale initialization at the beginning, similar to the monocular tightly coupled VIO \cite{QinTRO2018}. We get the necessary initial values by loosely aligning IMU preintegration with the vision-only structure. The estimator initialization for the multi-camera system consists of principal camera selection, vision-only SfM on principal camera, gyroscope calibration, and camera state initialization.

\subsubsection{Principal Camera Selection}
The initialization starts with a principal camera selection. In the sliding window, we check the feature correspondences in each camera between the latest frame and all previous frames. After we find stable feature tracking (more than 30 tracked features) between the latest frame and one typical frame in the sliding window, the camera with the most visual features is selected as the candidate of the principal camera. If the parallax is sufficient (more than 20 pixels), it is regarded as the principal camera. Otherwise, we continue the selection within other cameras. Once the principal camera is identified, the other cameras are characterized as subordinate cameras, and their initialization depends on the initialization of the principal camera.

\subsubsection{Vision-only SfM on Principal Camera}
Once we identify a principal camera and the corresponding two frames, we perform a vision-only SfM \cite{QinTRO2018} on the principal camera. We set the first principal camera frame $(\cdot)^{c_{p0}}$ as the reference world frame for SfM. Here we denote $c_p$ as the principal camera and $c_i$ as subordinate cameras. Given the extrinsic parameters $(\boldsymbol{p}^b_{c_i},\boldsymbol{q}^b_{c_i})$ between the camera and the IMU, we can recover the frame rotations of subordinate cameras with respect to $(\cdot)^{c_{p0}}$ as
\begin{equation}
\begin{aligned}
\boldsymbol{q}_{c_{ik}}^{c_{p0}} = \boldsymbol{q}_{c_{pk}}^{c_{p0}} \otimes (\boldsymbol{q}_{c_{p}}^{b})^{-1} \otimes \boldsymbol{q}_{c_{i}}^{b}  
\label{subordinate}
\end{aligned}
\end{equation}
% \\ \boldsymbol{p}_{c_{ik}}^{c_{p0}} &\approx \boldsymbol{p}_{c_{pk}}^{c_{p0}}  
% $$ \boldsymbol{p}_{c_{ik}}^{c_{p0}} = \boldsymbol{R}_{c_{pk}}^{c_{p0}} \boldsymbol{R}_{c_{p}}^{b}(\boldsymbol{p}_{c_{k}}^{b}-\boldsymbol{p}_{c_{p}}^{b}) +\boldsymbol{p}_{c_{pk}}^{c_{p0}}  \eqno(1) $$
Note that the scale of principal camera $s_p$ is unknown at this stage, so we cannot directly obtain the translation of subordinate cameras using extrinsic parameters. $s_p$ will be solved in the next.
% However, in a light-weight VIO system, the cameras and IMU are usually close to each other, so it is valid to use the translation of principal camera to approximate that of subordinate cameras. Secondly, the negative impact caused by subordinate camera positions is not critical and can be mitigated after the optimization.

% We can translate poses from the principal camera frame to body (IMU) frame as
% \begin{equation}
% \begin{aligned}
% \boldsymbol{q}_{b_{k}}^{c_{p0}} &= \boldsymbol{q}_{c_{pk}}^{c_{p0}} \otimes (\boldsymbol{q}_{c_{p}}^{b})^{-1} \\&= \boldsymbol{q}_{c_{ik}}^{c_{p0}} \otimes (\boldsymbol{q}_{c_{i}}^{b})^{-1} \\
%  s_p\boldsymbol{\overline{p}}_{b_{k}}^{c_{p0}} &= s_p\boldsymbol{\overline{p}}_{c_{pk}}^{c_{p0}}-\boldsymbol{R}_{b_k}^{c^0}\boldsymbol{p}_{c_{p}}^{b}
% \end{aligned}
% \end{equation}
% where $s_p$ is the unknown scaling parameter, which will be solved in the next.

% \subsubsection{Visual-Inertial Alignment}
\subsubsection{Gyroscope Bias Calibration}
Considering two consecutive frames $b_k$ and $b_{k+1}$ in the window, we get the rotation $\boldsymbol{q}^{c_{p0}}_{b_k}$, and  $\boldsymbol{q}^{c_{p0}}_{b_{k+1}}$. In addition, we have different values for each $\boldsymbol{q}^{c_{p0}}_{b_k}$, denoted as $\boldsymbol{q}^{c_{i0}}_{b_{ik}}$, where $i$ indicates that the body rotation is obtained from the \textit{i}-th camera using extrinsic parameter $(\boldsymbol{p}^b_{c_i},\boldsymbol{q}^b_{c_i})$. 
% Together with the relative constraint $\hat{\gamma}_{b_{k+1}}^{b_k}$, 
We linearize the IMU preintegration term with respect to gyroscope bias and minimize the following cost function:

\begin{equation}
\begin{aligned}
& \underset{\delta \boldsymbol{b}_w}{\text{min}} \underset{i \in \mathcal{C}}{\sum} \underset{j \in \mathcal{B}}{\sum} \Vert (\boldsymbol{q}_{c_{i(k+1)}}^{c_{p0}})^{-1} \otimes \boldsymbol{q}_{b_{ik}}^{c_{p0}} \otimes \gamma_{b_{k+1}}^{b_k} \Vert ^2
\end{aligned}
\end{equation}
% \\& \gamma_{b_{k+1}}^{b_k} \approx \hat{\gamma}_{b_{k+1}}^{b_k} \otimes  \begin{bmatrix}
% 1\\\frac{1}{2}\boldsymbol{J}^{\gamma}_{b_w}\delta \boldsymbol{b}_wa \end{bmatrix}
where $\mathcal{B}$, $\mathcal{C}$ index all frames in the window and all the cameras, respectively. $\gamma_{b_{k+1}}^{b_k}$ is the IMU preintegration constraint defined in \cite{QinTRO2018}.
%$\boldsymbol{J}^{\gamma}_{b_w}$ is the jacobian of $\gamma$ with respect to $\boldsymbol{b}_w$. 
In such a way, we get an initial calibration of the gyroscope bias $\boldsymbol{b}_w$. Then we follow the same way as in \cite{QinTRO2018} to repropagate all IMU preintegration terms using the new gyroscope bias.

\subsubsection{Camera States Initialization}
After the gyroscope bias is initialized, we move on to initialize velocity $\boldsymbol{v}^{b_k}_{b_k}$ in the body frame, gravity vector $\boldsymbol{g}^{c_p}$ and metric scale $s_p$ by aligning the up-to-scale visual structure with IMU measurements and refining the gravity \cite{QinTRO2018}.
% \begin{equation}
%     \mathcal{X}_p = [\boldsymbol{v}^{b_0}_{b_0},\boldsymbol{v}^{b_1}_{b_1},...,\boldsymbol{v}^{b_n}_{b_n},\boldsymbol{g}^{c_{p0}}, s_p]
% \end{equation}
% where $\boldsymbol{v}^{b_k}_{b_k}$ is velocity in the body frame while taking the \textit{k}th image, $\boldsymbol{g}^{c_p}$ is the gravity vector in the $c_{p0}$ frame, and $s_p$ scales the monocular SfM to metric units of principal camera.

% \subsubsection{Initialization of Subordinate Cameras}
After the visual alignment on the principal camera, we move on to initialize the subordinate cameras. We first proceed to recover the translation of subordinate cameras using the extrinsic between cameras and IMU
\begin{equation}
\begin{aligned}
\boldsymbol{p}_{c_{ik}}^{c_{p0}} = \boldsymbol{R}_{c_{pk}}^{c_{p0}} \boldsymbol{R}_{c_{p}}^{b}(\boldsymbol{p}_{c_{k}}^{b}-\boldsymbol{p}_{c_{p}}^{b}) +s_p\overline{\boldsymbol{p}}_{c_{pk}}^{c_{p0}}
\end{aligned}
\end{equation}
Now we triangulate all the feature points observed in the subordinate cameras. 
% \subsubsection{Completing Initialization}

Once the principal camera initializes successfully, the initialization is regarded as a success. All the values are rotated to the world frame $(\cdot)^w$ and fed into the tightly coupled multi-camera VIO.
% The velocity is known parameter since it is obtained in the previous step. The only unknown variables are 
% \begin{equation}
%     \mathcal{X}_i = [\boldsymbol{g}^{c_i}, s_i]
% \end{equation}
% Therefore, the linear least-squre problem remained to solve is 
% \begin{equation}
%     Min
% \end{equation}
% where 
% \begin{equation}
%     \boldsymbol{H} = []
% \end{equation}

% \subsection{Gravity Refinement and completing Initialization}
% We follow the same procedure in\cite{QinTRO2018} to refine the gravity in each camera and finally complete initialization.

\subsection{Tightly Coupled Multi-camera VIO}
After estimator initialization, a sliding window based tightly coupled multi-camera VIO proceeds. We incorporate the inverse distances of features in multiple cameras into the optimization. The state vector is defined as
\begin{equation}
\begin{aligned}
 \mathcal{X} &= [\boldsymbol{x}_0,\boldsymbol{x}_1,...,\boldsymbol{x}_n,\boldsymbol{x}_{c_1}^{b},\boldsymbol{x}_{c_2}^{b},...,\boldsymbol{x}_{c_t}^{b},\boldsymbol{\lambda}]\\\
 \boldsymbol{\lambda} &=[ \lambda_{c_1,0},...\lambda_{c_1,m_{c_1}},...,\lambda_{c_2,0},...\lambda_{c_t,m_{c_t}}]\\
\boldsymbol{x}_k &= [\boldsymbol{p}^{w}_{b_k},\boldsymbol{v}^{w}_{b_k},\boldsymbol{q}^{w}_{b_k},\boldsymbol{b}_{a},\boldsymbol{b}_{g}]\\
\boldsymbol{x}_{c_t}^{b} &= [\boldsymbol{p}^{b}_{c_t},\boldsymbol{q}^{b}_{c_t}] \label{variables}
\end{aligned}
\end{equation}
where $\boldsymbol{x}_k$ is the IMU state at the time when the \textit{k}-th image
is captured. It contains position, velocity, and orientation of
the IMU in the world frame, acceleration bias $\boldsymbol{b}_{a}$ and gyroscope bias $\boldsymbol{b}_{g}$ in the IMU body frame. $n$ is the total number of keyframes, $t$ is the total number of initialized cameras, and $m_i$ is the total number of features of the \textit{i}-th camera in the sliding window. $\lambda_{i,j}$ is the inverse distance of the \textit{j}-th feature in the \textit{i}-th camera from its first observation.

A visual-inertial bundle adjustment is performed to estimate the states. The formulation is
\begin{equation}
\begin{aligned}
 \underset{\mathcal{X}}{\text{min}} \bigg{\{}\Vert \boldsymbol{r}_p-\boldsymbol{H}_p \mathcal{X}\Vert^2
+  \underset{k \in \mathcal{K}}{\sum} \Vert \boldsymbol{r}_\mathcal{K}(\hat{\boldsymbol{z}}_{b_k}^{b_{k+1}}, \mathcal{X})  \Vert^2_{\boldsymbol{p}^{b_k}_{b_{k+1}}}
\\ + \underset{i \in \mathcal{C}}{\sum} \underset{(l,j) \in \mathcal{B}_i}\sum \alpha_i\rho(\Vert \boldsymbol{r}_{\mathcal{B}_i}(\hat{\boldsymbol{z}}_{l}^{c_{ij}}, \mathcal{X}) \Vert ^2_{\boldsymbol{p}^{c_{ij}}_{l}})
\bigg{\}}
\label{BA}
\end{aligned}
\end{equation}
 $\rho(s)$ is the cauchy robust function used to suppress outliers. $\mathcal{K}$ is the set of all IMU measurements. $\mathcal{C}$ is the set of all initialized cameras. $\mathcal{B}_i$ is the set of features that have been observed at least twice in the current sliding window in the \textit{i}-th camera.  $\{\boldsymbol{r}_p , \boldsymbol{H}_p\}$ is the prior information from marginalization. $\boldsymbol{r}_\mathcal{K}(\hat{\boldsymbol{z}}_{b_k}^{b_{k+1}}, \mathcal{X})$ is the residual for IMU. $\boldsymbol{r}_{\mathcal{B}_i}(\hat{\boldsymbol{z}}_{l}^{c_{ij}}, \mathcal{X})$ is the residual for visual measurements. The detailed definitions are presented in \cite{QinTRO2018}. $\alpha_i$ is defined as
 
 \begin{equation}
 \alpha_i = (\frac{N_i}{N_{max}})^2 
 \end{equation}

where $N_i$, $N_{max}$ are the number of features in the \textit{i}-th camera and the maximum number of features observed in one typical camera, respectively.
 
\subsection{Relocalization}
The relocalization generally follows the scheme in \cite{QinTRO2018}. The only difference is that the frames from different cameras are all stored in the previous map for loop detection and feature retrieval. A loop match detected in any camera will trigger the relocalization. 

\section{Experimental results}
In this section, we present the results of the VPI-enhanced VINS-Mono and the final multi-camera VIO.

\subsection{VINS-Mono with VPI}
To demonstrate the effectiveness of the proposed new front-end in terms of CPU usage and computation latency, we test different VIO methods on EuRoC dataset \cite{Burri25012016}. The used CPU for testing is \textit{AMD Ryzen 5800H with Radeon Graphics 3.20GHz}. For each video sequence, we conduct five trials for both the original VINS-Mono and our VPI enhanced VINS-Mono with Harris corner (VINS-VPI-Harris). In addition, we deploy another GPU enhanced front-end from VILIB \cite{Nagy2020} into VINS-Mono for comparison. In VILIB, we test two feature detectors, Harris and FAST \cite{rosten2008faster}. Therefore, we add another two versions for comparison, VINS-VILIB-FAST and VINS-VILIB-Harris.  We use the modules provided in rpg\_trajectory\_repository \cite{Zhang18iros} to quantitatively evaluate and compare the results. The main comparison metrics consist of 1) the CPU usage and 2) the time of different VIO front-ends, and 3) the average absolute trajectory errors (RMSE) on translation and rotation of 5 times' trials. Table \ref{nofisheye} shows the CPU usage and time cost of different VIO front-ends on the EuRoC dataset. 
%Note that the feature tracker provided in VILIB fails to provide stable tracking, so we keep the original tracker in VINS-Mono implemented by OpenCV. 

% !!!!!!!!!!!!!!!Xavier
% We also test VINS-Mono with VPI on NVIDIA Xavier AGX. In the experiments, the VIO setup starts from the same position, travels through the same trajectory and finally comes back to the starting position. The testing result is shown in Fig. 5. 
%   \begin{figure}[thpb]
%       \centering
%       \includegraphics[scale=0.5]{figures/titlefigure.pdf}
%       \caption{Comparison of CPU usage (single core) of different versions of VINS-Mono on Xavier. The first group is the result obtained from original VINS-Mono. The feature type is Shi-tomasi feature in the original VINS-Mono. The second group is the result obtained from VINS-Mono with VPI. The table below each picture shows the CPU usage on the front-end (feature tracking) and back-end (optimization). The green lines in the pictures are the estimated trajectories. The white dots around them are tracked feature points cloud. The red dot indicate the start position of the estimation.}
%       \label{figurelabel}
%   \end{figure}
\begin{table}[h]
\begin{center}
\caption{{\upshape Front-end CPU usage and feature extraction time \\ for different VIO methods on EuRoC dataset}}
\label{nofisheye}
\begin{tabular}{c|cc}
\hline
Method & CPU usage (\%) & Time (ms) \\
\hline
VINS-Mono & 52 & 15.4 \\
VINS-VPI-Harris & \textbf{31} &  \textbf{7.8} \\
\makecell[c]{ VINS-VILIB-FAST} & 49 & 8.5\\
\makecell[c]{VINS-VILIB-Harris} & 51 & 8.7\\
\hline
\end{tabular}
\vspace{-0.5cm}
\end{center}
\end{table}

Table \ref{nofisheye} shows VINS-Mono-VPI has the most significant reduction on CPU usage and time cost, achieving a reduction of 40.4\% and 50.6\% compared with the original VINS-Mono, respectively. Modules from VILIB (including VINS-VILIB-FAST and VINS-VILIB-Harris) reduce the front-end time cost, but they still have very high CPU usages. These results are from all the 11 video sequences in EuRoC dataset.

\begin{figure}[thpb]
      \centering
      \includegraphics[scale=0.6]{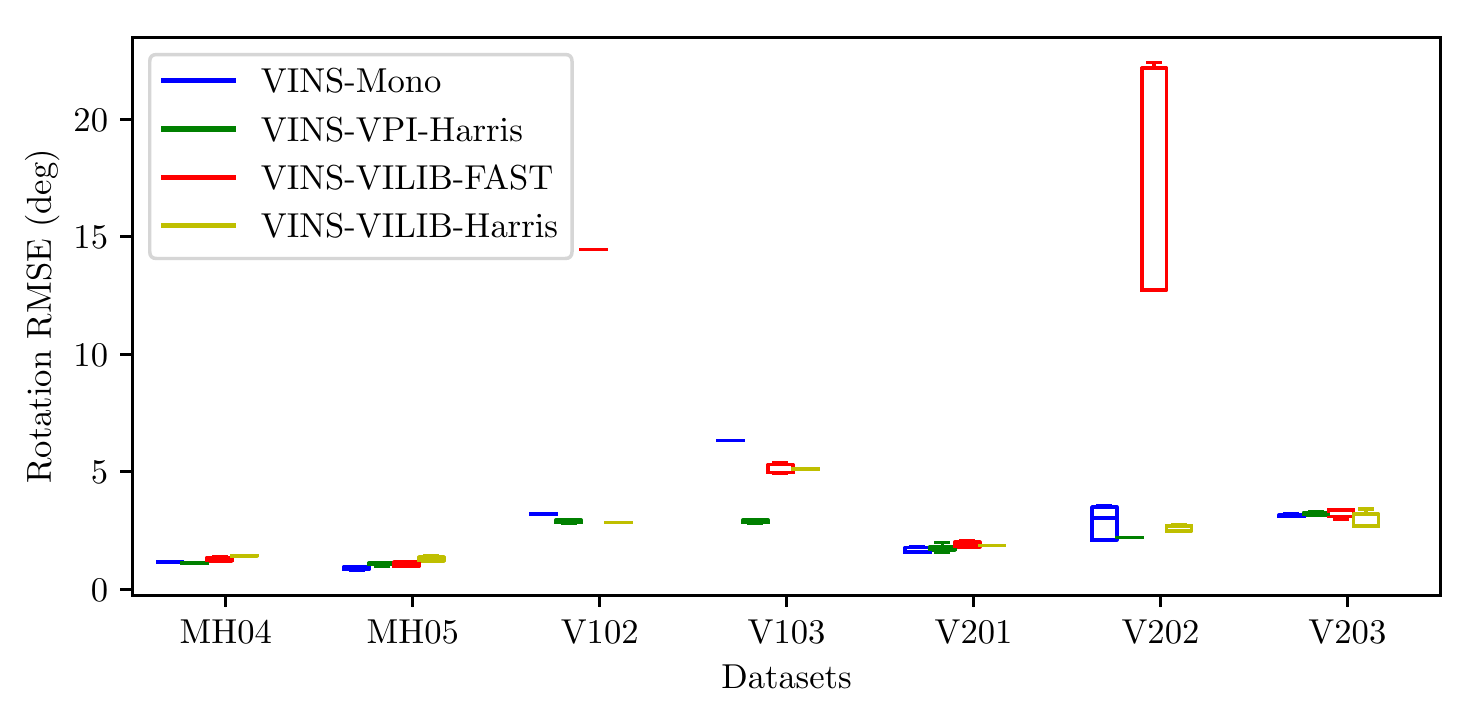}
            \caption{Rotation RMSE for different VIO methods.}
      \label{rrmse}
      \vspace{-0.4cm}
\end{figure}
   
\begin{figure}[thpb]
      \centering
      \includegraphics[scale=0.6]{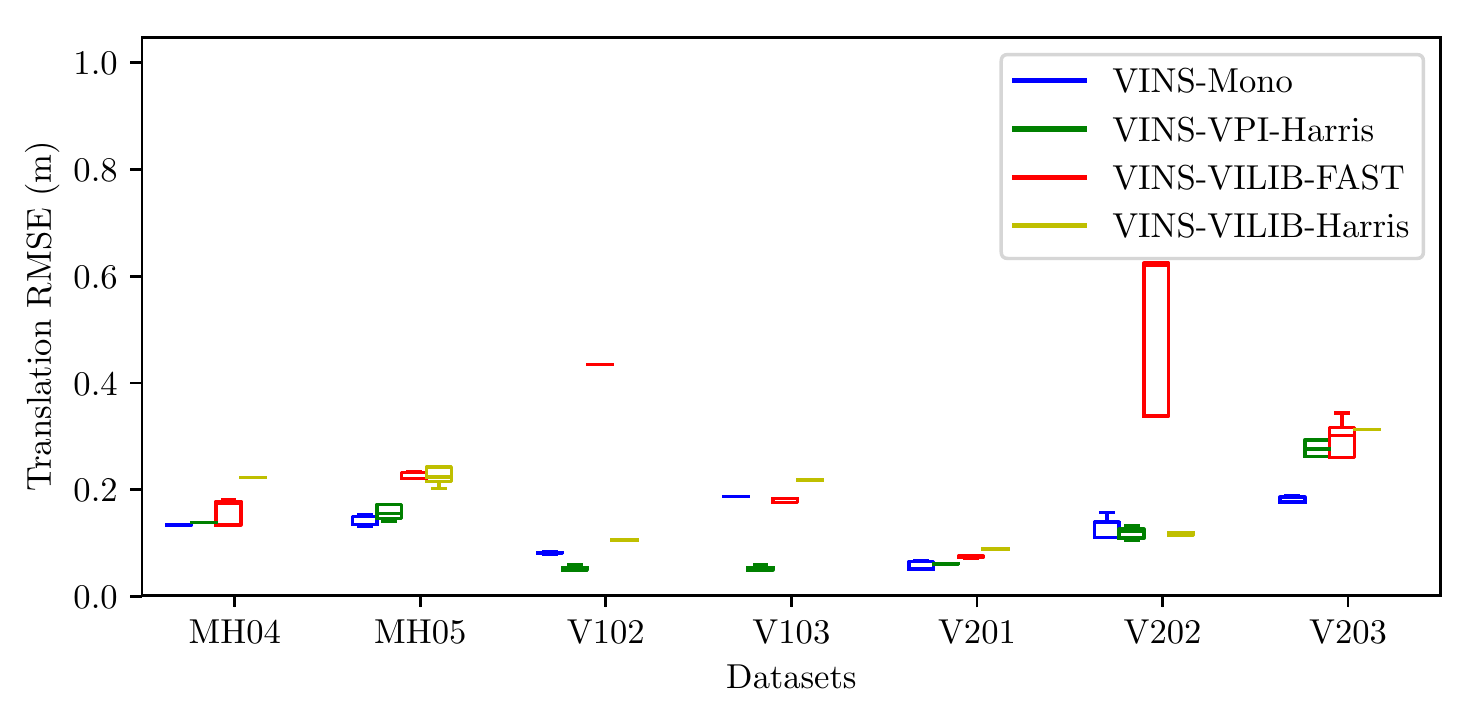}
      \caption{Translation RMSE for different VIO methods.}
      \label{trmse}
      \vspace{-0.3cm}
\end{figure}
 Fig. \ref{rrmse} and Fig. \ref{trmse} show the trajectory evaluation results on 7 relatively difficult sequences of EuRoC dataset. Compared with the original VINS-Mono, VINS-VPI-Harris has the most competitive performance in terms of translation and rotation errors. In some sequences, like V1\_02, V1\_03, and V2\_02, VINS-VPI-Harris obtains better accuracy.  On the other hand, VINS-Mono with VILIB loses accuracy to some extent, especially for the FAST feature when testing on V1\_02 and V2\_02 sequences. These results demonstrate that the VPI-enhanced VINS-Mono can not only reduce the CPU usage and time cost, but also maintain stable state estimation.
   
\subsection{Multi-camera VIO}
\begin{figure}[thpb]
    \centering
    \includegraphics[scale=0.3]{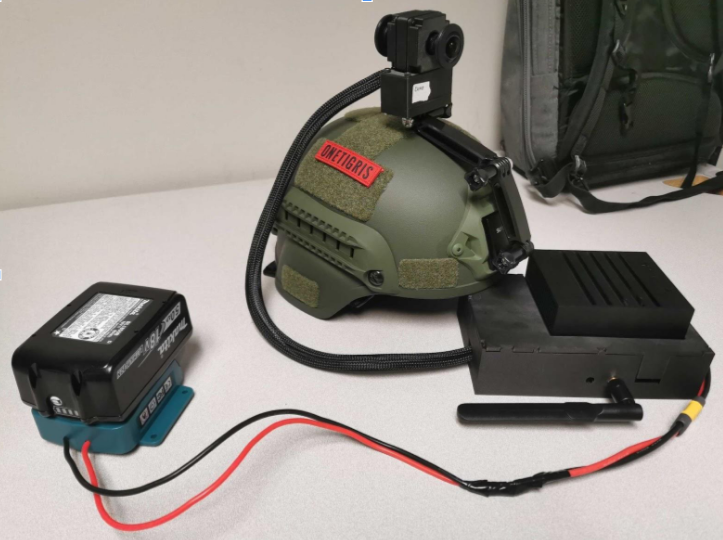}
    \caption{Our VIO setup. From left to right, the devices are: battery, helmet with fisheye cameras, and Xavier. Camera 1 and camera 2 are the cameras facing in the front and back, respectively.}
    \label{setup}
    \vspace{-0.3cm}
\end{figure}

To test the performance of the proposed multi-camera VIO method with the VPI-enhanced front-end and the tightly-coupled back-end, we collect our own data using a pedestrian carried system  (as shown Fig. \ref{setup}) with two non-overlapping fisheye cameras. The two Leopard IMX264 cameras capture fisheye images ($816\times 686 $ pixels images at 24.5Hz) with a synchronized Epson g365 IMU (200Hz).  We provide five datasets: Outdoors, Indoors, Corridor, NSH wall, and Smith Hall, to evaluate the performance and compare it with monocular VIO results. The outdoor and indoor datasets are mainly used to test the VIO system initialization performance. Since it is hard to obtain the ground truth trajectories for both indoor and outdoor datasets, the other three datasets are provided with accurate 3D point cloud maps \cite{yu2020monocular} for VIO result evaluation.  
% TODO
% add description

\begin{table}[h]
\setlength{\abovecaptionskip}{-0.25cm}
\caption{{\upshape Front-end CPU usage and feature extraction time \\
for different VIO methods with a fisheye camera}}
\label{fisheye}
% \label{table_example}
\begin{center}
\begin{tabular}{c|cc}
\hline
Method & \makecell[c]{CPU usage (\%) \\ (single core)} & \makecell[c]{Time (ms)\\(front-end)} \\
\hline
VINS-Mono & 91 & 20.6 \\
VINS-VPI-Harris & 43 &  11.7\\
VINS-VILIB-FAST & 82 & 28.7\\
VINS-VILIB-Harris & 91 & 16.2\\
\hline
\end{tabular}
\vspace{-0.4cm}
\end{center}
\end{table}

\begin{table}[h]
\setlength{\abovecaptionskip}{-0.25cm}
\caption{{\upshape CPU usage and computation time on optimization\\  of multi-camera VIO and VINS-Mono}}
\label{multiCPU}
\begin{center}
\begin{tabular}{c|cc}
\hline
VIO & \makecell[c]{VINS-Mono} & multi-camera VIO  \\
\hline
Time (back-end) & 40 ms & 40 ms \\
CPU usage (front-end) & 91 \% &43\%(each cam) \\
CPU usage (back-end) & 101\% &  105\%\\
\hline
\end{tabular}
\vspace{-0.4cm}
\end{center}
\end{table}

\begin{table*}[h]
\setlength{\abovecaptionskip}{-0.25cm}
\caption{{\upshape Analysis of the point clouds reconstructed using different VIO methods }}
\label{pcl}
\begin{center}
\begin{tabular}{c|cc|cc|cc|cc|cc}
\hline
\multirow{2}*{\diagbox{Scene}{Method}} & \multicolumn{2}{c|}{multi-camera VIO} & \multicolumn{2}{c|}{VINS-VPI-Harris cam-1} & \multicolumn{2}{c|}{VINS-VPI-Harris cam-2} & \multicolumn{2}{c|}{VINS-Mono cam-1} & \multicolumn{2}{c}{VINS-Mono cam-2} \\
\cline{2-11} &
 \textit{RMSE} & \textit{num} & \textit{RMSE} & \textit{num}& \textit{RMSE} & \textit{num}& \textit{RMSE} & \textit{num}& \textit{RMSE} & \textit{num}\\ 
\hline 
Corridor & \textcolor{blue}{0.105893} & \textcolor{red}{14743} & 0.111162 & 3017 & 0.114349 & 5909 & 0.114252 & 5410 & \textcolor{red}{0.104852} & \textcolor{blue}{10860} \\

Smith Hall & \textcolor{red}{0.110669} & \textcolor{red}{6751} & Fail & Fail & 0.116268 & 2771 & \textcolor{blue}{0.112309} & \textcolor{blue}{5843} & 0.112384 & 4642\\

NSH wall & \textcolor{blue}{0.113035} & \textcolor{red}{4523} & \textcolor{red}{0.112461} & 1514 & 0.120581 & 1506 & 0.116686 & \textcolor{blue}{3006} & 0.116389 & 1343\\
\hline
\end{tabular}
\vspace{-0.3cm}
\end{center}
\end{table*}

\begin{table*}[h]
\setlength{\abovecaptionskip}{-0.25cm}
\caption{{\upshape Initialization success rate and robustness of different VIO methods}}
\label{init}
\begin{center}
\begin{tabular}{c|ccccc}
\hline
\diagbox{Scene}{Method} & multi-camera VIO & VINS-Mono cam-1 & VINS-Mono cam-2 & VINS-VPI-Harris cam-1 & VINS-VPI-Harris cam-2 \\
\hline
Outdoor & \textbf{96.67\%} & 83.33\% & 66.67\% & 76.67\% & 63.33\% \\
Corridor & \textbf{98.57\%} & 25.71\% & 81.43\% & 28.57\% & 70.00\% \\
Indoor & \textbf{98.00\%} & 46.00\% & 62.00\% & 78.00\% & 94.00\% \\
NSH wall & \textbf{100.00\%} & \textbf{100.00\%} & \textbf{100.00\%} & 90.00\% & \textbf{100.00\%} \\
Smith Hall & \textbf{100.00}\% & 91.40\% & \textbf{100.00\%} & 85.70\% & 97.14\% \\
\hline
\end{tabular}
\vspace{-0.4cm}
\end{center}
\end{table*}

According to Table \ref{fisheye}, when the fisheye is used, the CPU usage and time cost increase compared with those on EuRoC dataset. This increase mainly comes from the larger image size and more feature points for undistortion. It is noticed that except VINS-VPI-Harris, all the other three methods have CPU usages close to 100\%. For VINS-Mono with FAST feature detector of VILIB, the time cost is even larger than that of VINS-Mono. Table \ref{multiCPU} compares the time cost and the CPU usage for the whole systems. We notice that the CPU usages on both front-end and back-end do not increase too much compared with VINS-Mono. Although multiple cameras introduce more features in the optimization, the computation time and CPU usage on optimization are roughly the same as that of VINS-Mono. Therefore, it further shows the potential of using the VPI-enhanced front-end on the multi-camera VIO system.

\begin{figure}[thpb]
      \centering
      \includegraphics[scale=0.18]{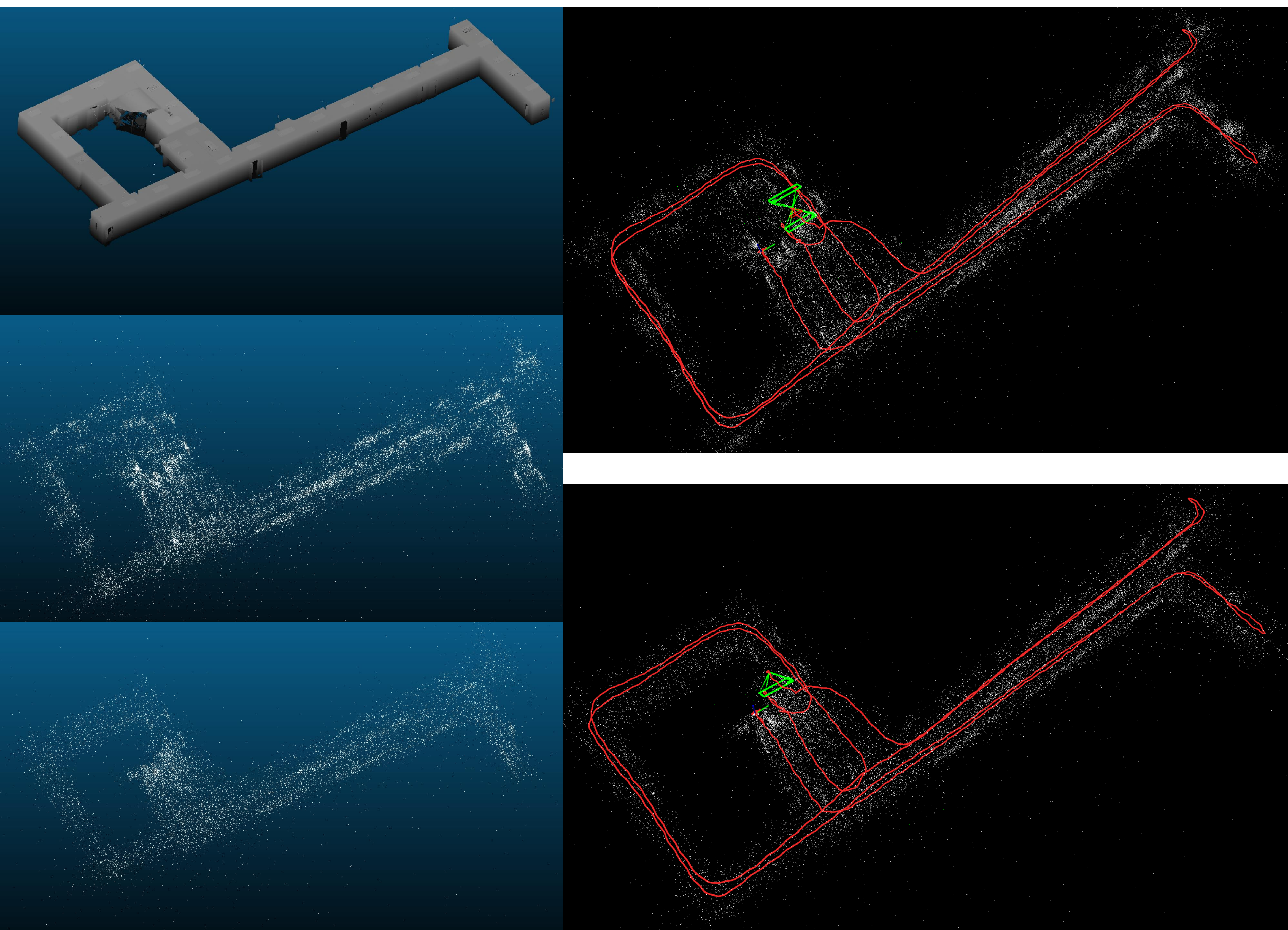}
      \caption{Point clouds and trajectories obtained in Corridor. \emph{Left (top-bottom)}: ground truth point clouds of Corridor, point clouds obtained by multi-camera VIO, point clouds obtained by VINS-Mono. \emph{Right (top-bottom)}: trajectory of multi-camera VIO, trajectory of VINS-Mono.}
      \label{cloud}
      \vspace{-0.4cm}
   \end{figure}

To qualitatively analyze the state estimation, the point clouds obtained by VIO are compared with the corresponding accurate 3D maps. We use the mean registration error ($RMSE$) and the number of registered points ($num$) as the metrics to measure the VIO performance. From Table \ref{pcl}, we can observe that the point clouds collected by multi-camera VIO have more registered points compared with other methods. With richer point clouds, the RMSE of multi-camera VIO is still competitive to the original VINS-mono. Fig. \ref{cloud} depicts an example of obtained point clouds and trajectories in Corridor. 
   
Furthermore, we use the initialization success rate as the criterion to evaluate the robustness. We uniformly select several time points on each dataset to initialize the VIO. If the VIO can initialize within a limited time (usually 4 seconds), it is regarded as a success. For each sequence, the experiment is repeated ten times for each VIO. The results are shown in Table \ref{init}. Overall, the multi-camera VIO has a remarkably higher initialization success rate compared with monocular VIO on one camera for all datasets. Especially for the Corridor dataset, the multi-camera setup improves the success rate significantly. This result shows that the multi-camera VIO is much more robust than monocular VIO on a single camera.

% \subsection{Discussion}
% VPI enhanced VINS-Mono both maintains the accuracy of VIO, and reduces the CPU usage and computational latency significantly. It reduces the CPU usage and computation latency by 40.4\% and 50.6\%, respectively. VILIB reduces the computation latency to some extend, but it does not reduce the CPU usage. It also has larger estimation errors compared to VINS-Mono and VINS-Mono with VPI. Therefore, we choose VPI to enhance the performance of the VIO, and use it to build the new front-end. 

\section{CONCLUSION}
In this paper, we first present a VPI enhanced front-end for visual-inertial odometry. With the efficient front-end, we propose a multi-camera VIO system with a tightly-coupled back-end. All the features observed in non-overlapping cameras are fed into the back-end for optimization. The VIO with VPI enhanced front-end reduces the CPU usage and time cost significantly, while maintains the state estimation accuracy. Without causing an extra burden on CPU and computation time, the multi-camera VIO system obtains higher robustness on system initialization and better state estimation than the monocular VIO.

In the future, we plan to incorporate the gstreamer pipeline into the multi-camera VIO. Gstreamer with VPI plugin can directly stream images from cameras to GPU and do the VIO front-end feature tracking without causing extra cost on CPU. Besides, we are also interested in enabling a brand new pose graph where scenes from different cameras can trigger the loop closure. 
%%%%%%%%%%%%%%%%%%%%%%%%%%%%%%%%%%%%%%%%%%%%%%%%%%%%%%%%%%%%%%%%%%%%%%%%%%%%%%%%
\newpage
% \section{ACKNOWLEDGEMENT}
% This work is supported by the Robotics Institute Summer
% Scholars Program (RISS), Carnegie Mellon University, the AirLab and
% the Chinese University of Hongkong, Shenzhen.

%%%%%%%%%%%%%%%%%%%%%%%%%%%%%%%%%%%%%%%%%%%%%%%%%%%%%%%%%%%%%%%%%%%%%%%%%%%%%%%%

\nocite{*}  % Without this, cite articles in text using \cite{...}
\bibliographystyle{IEEEtran}
\bibliography{./IEEEfull,refs}

\end{document}